# Incremental Nonparametric Weighted Feature Extraction for Online Subspace Pattern Classification

Hamid Abrishami Moghaddam
moghadam@eetd.kntu.ac.ir

Elaheh Raisi
elaheh@vt.edu

***Abstract*—In this paper, a new online method based on nonparametric weighted feature extraction (NWFE) is proposed. NWFE was introduced to enjoy optimum characteristics of linear discriminant analysis (LDA) and nonparametric discriminant analysis (NDA) while rectifying their drawbacks. It emphasizes the points near decision boundary by putting greater weights on them and deemphasizes other points. Incremental nonparametric weighted feature extraction (INWFE) is the online version of NWFE. INWFE has advantages of NWFE method such as extracting more than L-1 features in contrast to LDA. It is independent of the class distribution and performs well in complex distributed data. The effects of outliers are reduced due to the nature of its nonparametric scatter matrix. Furthermore, it is possible to add new samples asynchronously, i.e. whenever a new sample becomes available at any given time, it can be added to the algorithm. This is useful for many real world applications since all data cannot be available in advance. This method is implemented on Gaussian and non-Gaussian multidimensional data, a number of UCI datasets and Indian Pine dataset. Results are compared with NWFE in terms of classification accuracy and execution time. For nearest neighbour classifier it shows that this technique converges to NWFE at the end of learning process. In addition, the computational complexity is reduced in comparison with NWFE in terms of execution time.**

## I. INTRODUCTION

Curse of dimensionality has been known as a troubles issue in pattern recognition and computer vision techniques. Procedures that are analytically or computationally manageable in low dimensional spaces can become completely impractical in a space of 50 or 100 dimensions [1]. As a result, dimensionality reduction is considered to be one of the most important subjects in pattern classification. Reducing the dimension is used in different areas like image and signal processing, data transmission, data compression, etc. Two ways for dimensionality reduction are feature selection and feature extraction. In the former, we identify features which are unimportant in classification and discard them, i.e. we look for a d dimensional subspace out of the input p dimensional space (d<p). The latter entails finding a transformation from p features of the input space into a feature space with lower d dimension. This transformation could be a linear or nonlinear combination of the input variables. Besides, feature extraction is either supervised or unsupervised. In supervised, the goal is finding a transformation that maximizes a criterion for discrimination between classes. Among the supervised feature extraction methods, linear discriminant analysis (LDA) attracted lots of attention thanks to its special properties. LDA [1, 2] computes a transformation, which maximizes between class scatter while minimizes within class scatter; that way, the separability of classes in a lower dimension is maximized. However traditional LDA has some limitations:

- LDA suffers from small sample size (SSS) problem when the number of samples is lower than the dimension of data. It mostly happens in face recognition applications. In this case, $\mathbf{S}_w$ is ill-conditioned and becomes singular. To overcome this problem, many extensions of LDA have been introduced. One of them is regularized LDA [3, 4] in which a diagonal matrix αI is added to $\mathbf{S}_w$ ($\tilde{S}_w = S_w + \alpha I$) so that diagonal elements of within class scatter matrix is increased and singularity reduces. But there is no procedure to estimate the optimum value of parameter α. Another extension applies principal component analysis (PCA) as a pre-processing step and the output of PCA is used as input for LDA [5, 6, 7]. The drawback of this approach is that some useful discriminant information may be lost through PCA [8]. LDA/GSVD [9, 10] is another enhancement of LDA to avoid singularity by using Moore-Penrose generalized inverse. Nevertheless, GSVD requires that the whole data matrix be stored in main memory, so for large datasets the computational cost will be very high. LDA/QRD [11, 12] consists of two stages. The first stage maximizes between class scatter using QR decomposition and in the second stage LDA is done on small between and within class scatter matrices resulted from the previous stage. The other method is LDA using Pseudo-Inverse [12] which applies Pseudo-Inverse of $\mathbf{S}_w$ to avoid singularity.
- LDA can extract only L-1 features (L is the number of classes), because the rank of $\mathbf{S}_b$ matrix is at most L-1.
- LDA is optimal provided that the data distribution is Gaussian. If one or more classes be non-Gaussian in distribution, LDA may not work well because it depends on class means and covariance matrices [13]. Figure 1 depicted one of these situations.

Fukunaga [2] introduced a new method called nonparametric discriminant analysis (NDA) to rectify the disadvantages of LDA. In NDA, a nonparametric between class scatter matrix is defined which uses k-nearest neighbour for computing the local means. It exploits a weighting function so that the samples near the boundary decision will have greater weights. Therefore, the effects of outliers are decreased. NDA, itself has some drawbacks. There are some parameters which are usually decided by rules of thumb, and the within class scatter matrix is still parametric; thus, for small sample size problems it has singularity problem.

Nonparametric weighted feature extraction is proposed in [14]. It extends NDA by putting different weights on every sample for computing weighted means and defines new nonparametric between and within class scatter matrices.

All of the aforementioned methods assume that the whole data is available before the learning process begins. During the last decade, online learning has become important in some real world applications such as object tracking [15] and face recognition [16]; since it is impossible to know the exact number of samples beforehand. Online learning algorithms, which sequentially update eigenspace representation, act incrementally. Incremental versions of the methods mentioned previously were proposed. ILDA [17, 18, 19, 20], ILDA/GSVD [16], IDR/QRD [21] and incremental NDA [22] are the incremental forms of LDA, LDA/GSVD, LDA/QRD and NDA, respectively. Although these techniques have become online, their disadvantages persist.

In this paper, we propose incremental nonparametric weighted feature extraction (INWFE) to overcome the above drawbacks while performing sequentially. This algorithm is implemented on Gaussian and non-Gaussian multidimensional data, 8 datasets of UCI database [23] and Indian Pine dataset [24]. The results are compared with NWFE in terms of classification accuracy and execution time. We show that our approach converges to NWFE at the end of the learning process and the time required to recalculate the scatter matrices is less than NWFE.

The paper is organized as follows: In the following section, we briefly review nonparametric weighted feature extraction method. The next section presents the proposed method in details. Experimental results on Gaussian, non-Gaussian and some datasets of UCI will be given in section 4. Finally, we will conclude in the final section.

For convenience, we present in Table 1 the important notations and symbols used in the paper.

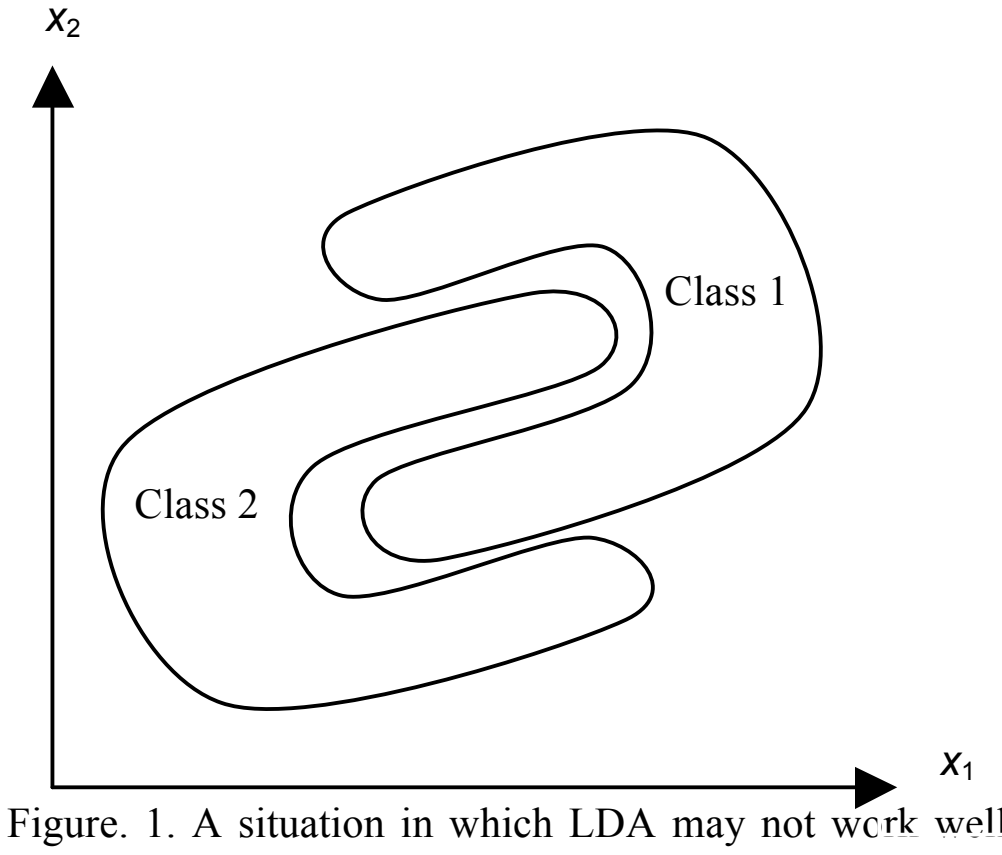

Figure. 1. A situation in which LDA may not work well since the within class covariance of matrices 1 and 2 do not reflect the actual scatter of the data

TABLE 1. Notations used in this paper.

| Notation | Description | Notation | Description |
|---|---|---|---|
| N | Number of training data points | L | The number of classes |
| $\mathbf{S}_w$ | Within class scatter matrix | $\mathbf{S}_b$ | Between class scatter matrix |
| $P_i$ | Prior probability of *i*-th class | $\mathbf{x}_l^i$ | *l*-th point of *i*-th class |
| d(a,b) | Euclidean distance between vectors of a and b | $N_i$ | Number of training data points in *i*-th class |
| $\mathbf{m}_j(\mathbf{x}_l^i)$ | The weighted mean of $\mathbf{x}_l^i$ in class j | $\lambda_l^{(i,j)}$ | weight between $\mathbf{x}_l^i$ and $\mathbf{m}_j(\mathbf{x}_l^i)$ |

## II. Nonparametric Weighted Feature Extraction

In [14], a new feature extraction method called nonparametric weighted feature extraction is proposed to resolve problems of NDA. However, in NDA, some parameters are obtained by trial and error. Moreover, within class matrix is still parametric and for small sample size problems, NDA encounters singularity. For estimating the local mean and weightings, NDA uses only k nearest neighbors (kNN) to emphasize points close to the boundary. Since all kNN points do not carry the same amount of information to find class boundary, it may cause the loss of some information. Based on this concern, in NWFE, new nonparametric Sw and Sb are defined by computing the weighted mean for obtaining more than L-1 features. In [14], Sb is defined as:

$$S_b^{NW} = \sum_{i=1}^{L} P_i \sum_{\substack{j=1 \\ j\neq i}}^{L} \sum_{l=1}^{N_i} \frac{\lambda_l^{i,j}}{N_j} (\mathbf{x}_l^i - \mathbf{m}_j(\mathbf{x}_l^i))(\mathbf{x}_l^i - \mathbf{m}_j(\mathbf{x}_l^i)) \quad (1)$$

$N_i$ is training sample size of class i and $P_i$ denotes the prior probability of class i. Scatter matrix weight $\lambda_l^{(i,j)}$ is a function $\mathbf{x}_l^i$ of and $\mathbf{m}_j(\mathbf{x}_l^i)$

$$\lambda_l^{i,j} = \frac{\text{dist}(\mathbf{x}_l^i, \mathbf{m}_j(\mathbf{x}_l^i))^{-1}}{\sum_{t=1}^{N_i} \text{dist}(\mathbf{x}_t^i, \mathbf{m}_j(\mathbf{x}_t^i))^{-1}} \quad (2)$$

dist(a,b) denotes the Euclidean distance from **a** to **b**. If the distance between $\mathbf{x}_l^i$ and $\mathbf{m}_j(\mathbf{x}_l^i)$ is small, $\lambda_l^{(i,j)}$ will be close to 1; otherwise, $\lambda_l^{(i,j)}$ will be close to 0. The sum of the $\lambda_l^{(i,j)}$ for class i is 1. $\mathbf{m}_j(\mathbf{x}_l^i)$ denotes the weighted mean of in class j and is defined as:

$$\mathbf{m}_j(\mathbf{x}_l^i) = \sum_{k=1}^{N_j} w_{lk}^{i,j} \mathbf{x}_k^j \quad (3)$$

Where

$$w_{lk}^{i,j} = \frac{\text{dist}(\mathbf{x}_l^i, \mathbf{x}_k^j)^{-1}}{\sum_{t=1}^{N_i} \text{dist}(\mathbf{x}_t^i, \mathbf{x}_k^j)^{-1}} \quad (4)$$

The nonparametric within class scatter matrix is defined by:

$$S_w^{NW} = \sum_{i=1}^{L} P_i \sum_{l=1}^{N_i} \frac{\lambda_l^{i,j}}{N_i} (\mathbf{x}_l^i - \mathbf{m}_i(\mathbf{x}_l^i))(\mathbf{x}_l^i - \mathbf{m}_i(\mathbf{x}_l^i))^T \quad (5)$$

It is worthwhile to note that in equations (1) and (5), $P_i / N_i = 1/N$ represents a constant coefficient and can be neglected. As proposed in [14], in order to prevent the singularity of within class scatter matrix, in this paper the following regularization technique is used

$$\mathbf{S}_w^{NW} = 0.5\mathbf{S}_w^{NW} + 0.5\text{diag}(\mathbf{S}_w^{NW}) \quad (6)$$

where diag($\mathbf{S}_w^{NW}$) is the diagonal parts of matrix $\mathbf{S}_w^{NW}$. The NWFE algorithm is as follows:

(1) The distance between each pair of sample points is computed and the distance matrix is calculated,

(2) $w_{lk}^{i,j}$ is computed using distance matrix,

(3) Weighted means $\mathbf{m}_j(\mathbf{x}_l^i)$ are calculated using $w_{lk}^{i,j}$,

(4) Scatter matrix weight $\lambda_l^{(i,j)}$ is computed,

(5) $\mathbf{S}_b^{NW}$ and $\mathbf{S}_w^{NW}$ are calculated,

(6) Regularize $\mathbf{S}_w^{NW}$,

(7) Features are extracted using LDA method using new scatter matrices $\mathbf{S}_b^{NW}$, $\mathbf{S}_w^{NW}$

## III. Incremental Nonparametric Weighted Feature Extraction

NWFE was proposed to improve NDA, however it was a batch mode algorithm, i.e. all the samples have to be available in advance. As we know, this is impossible for most of the real world applications. Here, we propose incremental NWFE in which training points are added sequentially. Similar to NDA, Sw and Sb are calculated from at least two classes. When a new training point y arrives, we have two situations.

### *A. The new training pattern belongs to an existing class*

Let's assume the new pattern **y** belongs to one of the existing classes, $C_E$, $1 \le E \le L$, some values of $w_{lk}^{i,j}$, $\mathbf{m}_j(\mathbf{x}_l^i)$ and $\lambda_l^{(i,j)}$ will change and some new $w_{lk}^{i,j}$, $\mathbf{m}_j(\mathbf{x}_l^i)$ and $\lambda_l^{(i,j)}$ will be added.

- Based on the definition of $w_{lk}^{i,j}$, every $w_{lk}^{i,j}$ depends on all patterns of *i*-th class, so $w_{lk}^{i,j}$ for i = E will be changed and new weights $w_{lk}^{E,j}$ , $w_{lk}^{i,E}$ , i,j=1,…,L $w_{lk}^{i,E}$ i,j=1,…,L for the new pattern is calculated,
- $\mathbf{m}_j(\mathbf{x}_l^i)$ depends on all patterns of *j*-th class; thus $\mathbf{m}_j(\mathbf{x}_l^i)$ for j=E change and $\mathbf{m}_j(\mathbf{x}_y^E)$ for j=1,…,L, j≠E will be added.
- $\lambda_l^{(i,j)}$ depends on all patterns of *i*-th class, so $\lambda_l^{(E,j)}$, $\lambda_l^{(i,E)}$ will be recalculated and $\lambda_y^{(E,j)}$ j=1,…,L , j≠E are added.
- As in [22], we use the same definitions for updating $\mathbf{S}_b$ and $\mathbf{S}_w$. Equation (7) is used to update $\mathbf{S}_b$.

$$\mathbf{S}_b' = \mathbf{S}_b - \mathbf{S}_b^{in}(C_E) + \mathbf{S}_b^{in}(C_{E'}) + \mathbf{S}_b^{out}(\mathbf{y}^{C_E}) \quad (7)$$

Where $C_E' = C_E \cup \{\mathbf{y}^{C_E}\}$ . **y** is the new pattern and E is the class of **y**. $\mathbf{S}_b^{in}(C_E)$ is the covariance matrix between the existing classes and the class which is about to change. $\mathbf{S}_b^{in}(C_{E'})$ is the covariance matrix between existing classes and the updated class $C_E'$ . $\mathbf{S}_b^{out}(\mathbf{y}^{C_E})$ represents the covariance matrix between vector $\mathbf{y}^{C_E}$ and the other classes. $\mathbf{S}_b^{in}(C_E)$ is calculated by the following equation:

$$\mathbf{S}_b^{in}(C_E) = \sum_{j=1,j\neq E}^{L}\sum_{l=1}^{N_E}\lambda_l^{E,j}(\mathbf{x}_l^E - \mathbf{m}_j(\mathbf{x}_l^E))(\mathbf{x}_l^E - \mathbf{m}_j(\mathbf{x}_l^E))^T$$

$$+ \sum_{i=1,i\neq E}^{L}\sum_{l=1}^{N_i}\lambda_l^{i,E}(\mathbf{x}_l^i - \mathbf{m}_E(\mathbf{x}_l^i))(\mathbf{x}_l^i - \mathbf{m}_E(\mathbf{x}_l^i))^T \qquad (8)$$

and $\mathbf{S}_b^{out}(\mathbf{y}^{C_E})$ is calculated by:

$$\mathbf{S}_b^{out}(\mathbf{y}^{C_E}) = \sum_{j=1,j\neq E}^{L}\lambda_y^{E,j}(\mathbf{y}^E - \mathbf{m}_j(\mathbf{y}^E))(\mathbf{y}^E - \mathbf{m}_j(\mathbf{y}^E))^T \qquad (9)$$

And:

$$\mathbf{S}_W' = \mathbf{S}_W - \mathbf{S}_W(C_E) + \mathbf{S}_W(C_{E'}) \qquad (10)$$

where $\mathbf{S}_w(C_E)$ is the within class covariance of class E before adding the new pattern and $\mathbf{S}_w(C_E')$ is the within class covariance of class E, after adding the new pattern. $\mathbf{S}_w(C_E)$ is defined by:

$$\mathbf{S}_w(C_E) = \sum_{l=1}^{N_C}\lambda_l^{C,C}(\mathbf{x}_l^C - \mathbf{m}_C(\mathbf{x}_l^C))(\mathbf{x}_l^C - \mathbf{m}_C(\mathbf{x}_l^C))^T \qquad (11)$$

Figure 2 shows this situation. New pattern is represented by the solid red 4-point star which belongs to the class $C_E$. $C_E'$ is the class after adding the new pattern **y**.

*B. The new training pattern belongs to a new class*

In this situation, new pattern belongs to a new class $C_{L+1}$. Figure 3 depicts this situation. The links show the old and new status with the class before and after the introduction of the new point. New pattern is addressed by solid red 5-point star. In this way, some new values of **W**, **m** and λ will be added.

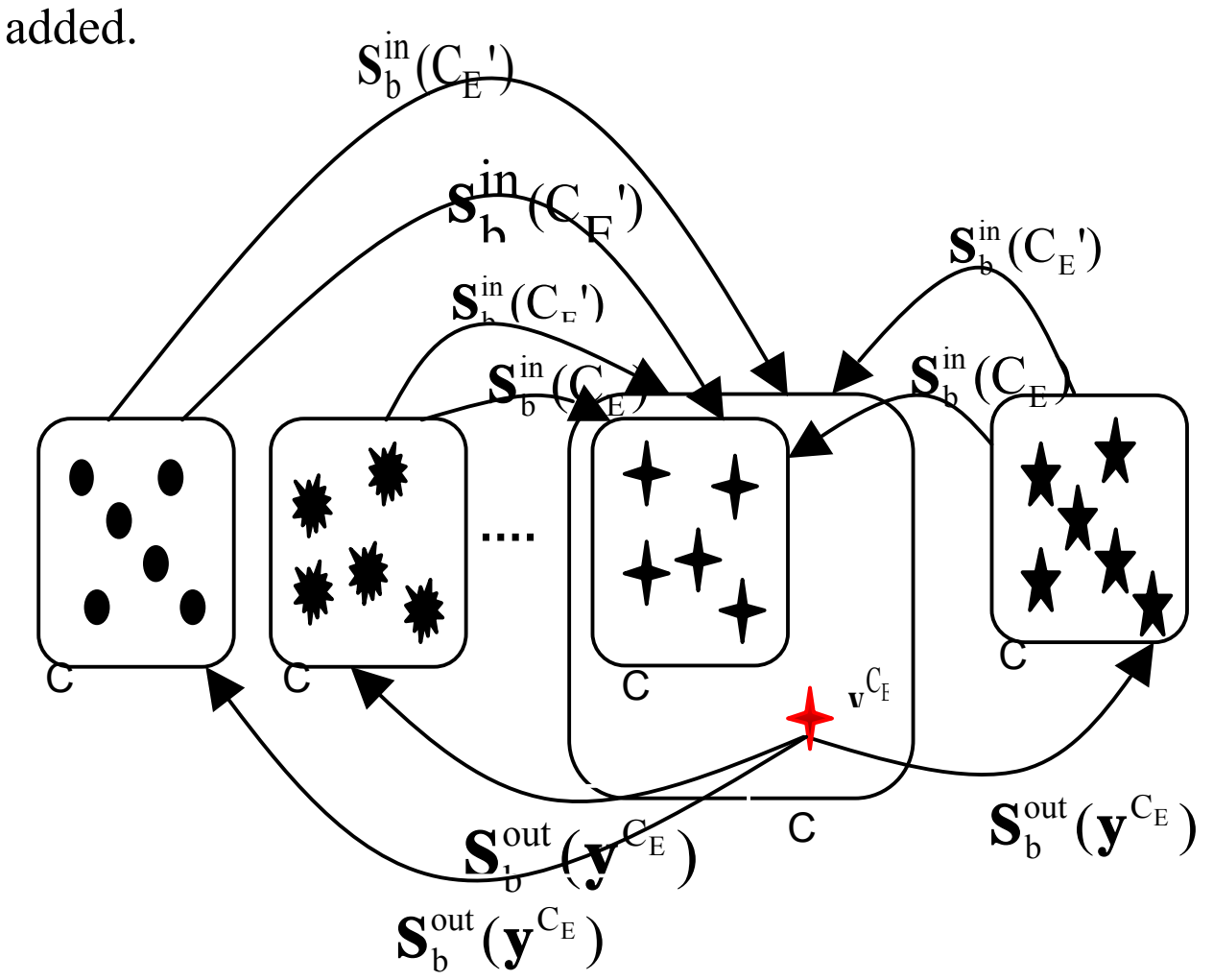

Figure 2. The new pattern belongs to an existing class

- $w_{lk}^{i,j}$ for i=L+1,j=1,…,L, and j=L+1,i=1,…,L,
- $\mathbf{m}_{L+1}(\mathbf{x}_l^i)$ $\mathbf{m}_{L+1}(\mathbf{x}_l^i)$i=1,…,L and j=L+1,i=1,…,L,
- (8) $\lambda_l^{(i,j)}$ for i=L+1,j=1,…,L and $\lambda_l^{(i,j)}$ for j=L+1,i=1,…,L.

The formula for recursive calculation of $\mathbf{S}_b$ is defined in equation (12)

$$\mathbf{S}_b' = \mathbf{S}_b - \mathbf{S}_b^{out}(C_{L+1}) + \mathbf{S}_b^{in}(C_{L+1}) \qquad (12)$$

Here $\mathbf{S}_b^{out}(C_{L+1})$ and $\mathbf{S}_b^{in}(C_{L+1})$ are given by:

$$\mathbf{S}_b^{out}(C_{L+1}) = \sum_{j=1}^{L}\lambda_y^{E,j}(\mathbf{y}^E - \mathbf{m}_j(\mathbf{y}^E))(\mathbf{y}^E - \mathbf{m}_j(\mathbf{y}^E))^T \qquad (13)$$

$$\mathbf{S}_b^{in}(C_{L+1}) = \sum_{i=1}^{L}\sum_{l=1}^{N_i}\lambda_l^{i,E}(\mathbf{x}_l^i - \mathbf{m}_E(\mathbf{x}_l^i))(\mathbf{x}_l^i - \mathbf{m}_E(\mathbf{x}_l^i))^T \qquad (14)$$

where $\mathbf{S}_W'$ matrix remains unchanged:

$$\mathbf{S}_w' = \mathbf{S}_w \qquad (15)$$

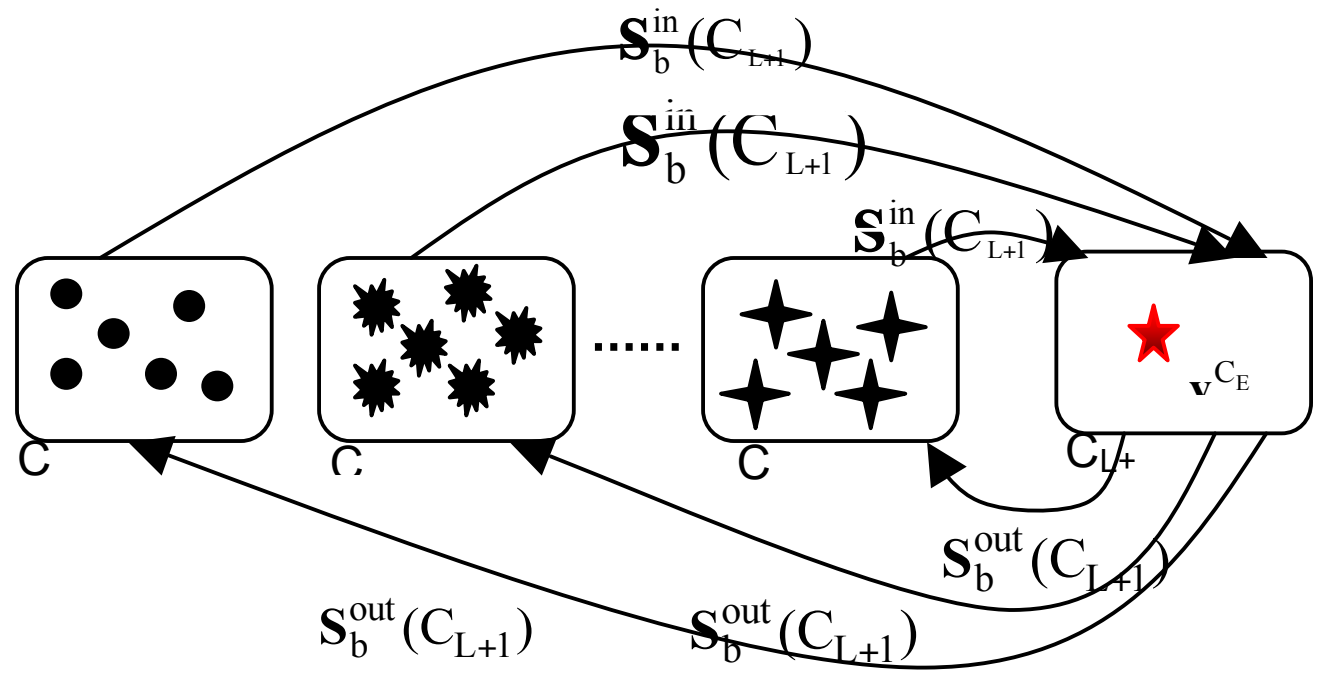

Figure 3. The new pattern belongs to a new class

## IV. EXPERIMENTS AND RESULTS

In this paper, the nearest neighbor is used for classification. When a test sample arrives, it is projected to the NWFE eigenspace and classified based on the class of its nearest neighbor. We use tenfold cross-validation technique i.e. the dataset is divided into 10 groups of equal size. The training-testing process is repeated ten times and each time one of these ten groups becomes test set while the remaining 90% of data is used as train set. In each experiment, about 20% of training data is used for initializing INWFE, and the rest is used as incremental data. INWFE is updated after each sample is presented. In these experiments, we applied synthetic and real world data. In the former, Gaussian and non-Gaussian data are generated. For generating non-Gaussian data, Gaussian Mixture Model technique is applied [1]. For the real data, we used some datasets of UCI database [23] and a remote sensing hyperspectral image of Indian Pine [24]. To overcome the singularity problem, the regularization technique is used. Table 2 shows the properties of 8 datasets from UCI and synthetic data parameters.

Table 2. Overview of evaluated UCI datasets and synthetic Gaussian and non-Gaussian data: S1 = sonar, S2 =liver, S3 = iris, S4 = wine, S5 = vehicle, S6 = glass, S7 = segmentation, S8 = vowel, G= Gaussian data, NG= non-Gaussian data

| Data Type | Index | No. classes | original dim | No. samples |
|---|---|---|---|---|
| | S1 | 2 | 60 | 208 |
| | S2 | 2 | 6 | 345 |
| | S3 | 3 | 4 | 150 |
| | S4 | 3 | 13 | 178 |
| UCI Datasets | S5 | 4 | 18 | 846 |
| | S6 | 6 | 9 | 214 |
| | S7 | 7 | 19 | 2310 |
| | S8 | 11 | 10 | 528 |
| Gaussian Data | G | 3 | 6 | 300 |
| Non-Gaussian Data | NG | 2 | 4 | 150 |

The datasets selected from UCI Machine Learning Repository ranges from small number of classes and a few samples per class to the large number of classes and many samples per class. Implementations of NWFE and INWFE were performed on these datasets.

For generating Gaussian data, we take three classes, 100 samples per class with six dimensions. On the other hand, for generating non-Gaussian data, we suppose two classes; each class consisting of three groups of 25 Gaussian data based on Gaussian mixture model technique; so the total number of data in this experiment is 300 for Gaussian and 150 for non-Gaussian distribution. The mean and covariance values for generating Gaussian and non-Gaussian data are show in Table3.

Regarding the high correlation between neighboring channels, we suppressed 92 channels by keeping one out of two neighboring channels and used the remaining 93 channels as experimental data. A color composite of the image and its ground truth map are shown in Figures 4.a and 4.b, respectively. In this study, in order to speed the training time, 7 smallest classes are removed and the size of 8 remaining classes is reduced to about 500 pixels. The number of labeled samples per class is given in Table 4.

There is no special rule for choosing the final dimension of NWFE. Here, for 6 datasets of UCI and two synthetic datasets, we evaluate the accuracy with respect to dimensionality from 1 to min(OD,15), where OD represents the original dimensionality (Figure 5). However, since Vehicle, Segmentation and Indian Pine datasets have large sample size, we had to select the NWFE dimension by rule of thumb for these databases.

Table 5 indicates the average recognition rate (for 10 iterations) and confidence intervals. This interval is calculated by:

$$\mathrm{var} = \frac{\mathrm{std(rec_rate)}}{\sqrt{\mathrm{max_iter}}} \qquad (16)$$

std(recog_rate) is standard deviation of recognition rates and max_iter = 10.

As mentioned, for 6 datasets of UCI and two synthetic datasets, different dimensionalities are tested and so different accuracies are resulted. In Table 5 the column labeled with NWFE-dim represents the dimension for which the best accuracy has been obtained.

As indicated in Table 5, at the end of learning process the performance of INWFE approaches to the performance of NWFE.

Figure 6 depicts the evolution of recognition rate as a function of input samples in small size Iris (Figure 6.a), large size Vowel (Figure 6.b) datasets of UCI and synthetic Gaussian (Figure 6.c) and non-Gaussian (Figure 6.d) data. As illustrated, INWFE achieves the same recognition rate as NWFE when all the data are introduced to the algorithm.

In NWFE, whenever a new sample is added, $\mathbf{S}_b$ and $\mathbf{S}_w$ matrices have to be recalculated from beginning using the whole data; however in INWFE, by adding new sample, $\mathbf{S}_b$ and $\mathbf{S}_w$ will be updated according to (7), (10) and (12) using their current values and the new sample data. In this study, we take the time required to recalculate $\mathbf{S}_b$ in NWFE and the time for updating $\mathbf{S}_b$ in INWFE method. The results for 6 datasets of UCI database and two Gaussian and non-Gaussian multidimensional data are presented in Table 6. The time required for recalculating $\mathbf{S}_b$ matrix is shown in seconds for every 100 samples. The experiment is performed in MATLAB on a 2.70 GHz Pentium Dual-Core computer, with 2 GB of memory. From Table 6, it can be seen that INWFE gives better results in terms of time complexity.

Table 3. Mean and covariance values for generating Gaussian and non-Gaussian data

| Data type | | | Mean | Covariance |
|---|---|---|---|---|
| Gaussian | Class1 | | [1 1 1 1 1 1] | $\sigma_1 = 1.5$ $\sigma_2 = .5$ $\sigma_3 = .3$ $\sigma_4 = 1$ $\sigma_5 = .3$ $\sigma_6 = .8$ |
| | Class2 | | [5 3 3 2 −1 2] | $\sigma_1 = 1$ $\sigma_2 = 1$ $\sigma_3 = .7$ $\sigma_4 = .5$ $\sigma_5 = 1$ $\sigma_6 = .6$ |
| | Class3 | | [3 2 2 1.5 3 .3] | $\sigma_1 = .5$ $\sigma_2 = 1$ $\sigma_3 = .5$ $\sigma_4 = 1$ $\sigma_5 = .6$ $\sigma_6 = .8$ |
| Non-Gaussian | Class1 | Group1 | [1 1 1 1] | $\sigma_1 = 1.5$ $\sigma_2 = .5$ $\sigma_3 = .5$ $\sigma_4 = .2$ |
| | | Group2 | [1.5 3 1.5 .3] | $\sigma_1 = 1$ $\sigma_2 = 1$ $\sigma_3 = 1$ $\sigma_4 = .6$ |
| | | Group3 | [3 2 1.7 .5] | $\sigma_1 = .5$ $\sigma_2 = .5$ $\sigma_3 = 1$ $\sigma_4 = 1$ |
| | Class2 | Group1 | [5 2 −1 1.7] | $\sigma_1 = .5$ $\sigma_2 = 1.5$ $\sigma_3 = 1$ $\sigma_4 = .2$ |
| | | Group2 | [4 2.8 3 3] | $\sigma_1 = 1$ $\sigma_2 = .6$ $\sigma_3 = .6$ $\sigma_4 = 1$ |
| | | Group3 | [4.6 3.5 −2 −3] | $\sigma_1 = .5$ $\sigma_2 = 1.5$ $\sigma_3 = 1$ $\sigma_4 = .3$ |

Table 4.The number of samples per class chosen from Indian Pine

| **Class Name** | **Size** |
|---|---|
| **Corn-no till** | **516** |
| **Corn-min** | **526** |
| **Grass/Trees** | **510** |
| **Hay-windrowed** | **454** |
| **Soybeans-no till** | **527** |
| **Soybeans-min** | **522** |
| **Soybeans-clean** | **505** |
| **Woods** | **504** |

Table 5. Comparison between NWFE and INWFE accuracy with NN classifier
S1 = sonar, S2 =liver, S3 = iris, S4 = wine, S5 = vehicle, S6 = glass, S7 = segmentation, S8 = vowel, G= Gaussian data, NG= non-Gaussian data, IP = Indian Pine.

| **Index** | **NWFE-dim** | **NWFE** | **INWFE** |
|---|---|---|---|
| S1 | 11 | 84.2875± 2.8439 | 86.1905 ± 2.4020 |
| S2 | 3 | 59.5042 ± 1.9472 | 61.5966± 1.7383 |
| S3 | 4 | 98.6667± 0.8889 | 97.3334± 1.0886 |
| S4 | 13 | 94.9306± 1.5426 | 95.4167± 1.6827 |
| S5 | 9 | 67.7081±1.6548 | 68.2123±1.5632 |
| S6 | 5 | 72.3619 ±3.0754 | 69.6571±3.0173 |
| S7 | 6 | 96.5412 ± 1.9860 | 95.1593 ± 1.3021 |
| S8 | 10 | 97.7063±0.7573 | 96.9515± 0.7747 |
| G | 5 | 91.3333 ±2.0154 | 92.3333 ±1.1966 |
| NG | 4 | 85.3333 ± 1.9373 | 84.6667± 2.0000 |
| IP | 10 | 94.3334±2.0521 | 93.6667± 2.5890 |

Table 6. Comparison between NWFE and INWFE time complexity in seconds for six datasets of UCI and two synthetic data. S1 = sonar, S2 =liver, S3 = iris, S4 = wine, S6 = glass, S8 = vowel, G= Gaussian data, NG= non-Gaussian data.

| Number of samples / Method | | 100 | 200 | 300 | 400 | 500 |
|---|---|---|---|---|---|---|
| **S1** | **NWFE** | **11.0310** | **63.0320** | | | |
| | **INWFE** | **7.9220** | **51.8440** | | | |
| **S2** | **NWFE** | **5.8120** | **46.9690** | **221.6720** | | |
| | **INWFE** | **15.4220** | **60.2350** | **163.6400** | | |
| **S3** | **NWFE** | **9.6250** | | | | |
| | **INWFE** | **8.3900** | | | | |
| **S4** | **NWFE** | **4.1720** | | | | |
| | **INWFE** | **4.062** | | | | |
| **S6** | **NWFE** | **3.3750** | **22.2500** | | | |
| | **INWFE** | **2.60920** | **6.9380** | | | |
| **S8** | **NWFE** | **1.4530** | **6.8600** | **31.3750** | **71.4680** | **345.4070** |
| | **INWFE** | **0.5470** | **5.9380** | **12.3590** | **44.5000** | **55.5620** |
| **G** | **NWFE** | **4.0310** | **32.7500** | **116.3590** | | |
| | **INWFE** | **2.2820** | **16.6870** | **63.9690** | | |
| **NG** | **NWFE** | **5.6250** | | | | |
| | **INWFE** | **4.2350** | | | | |

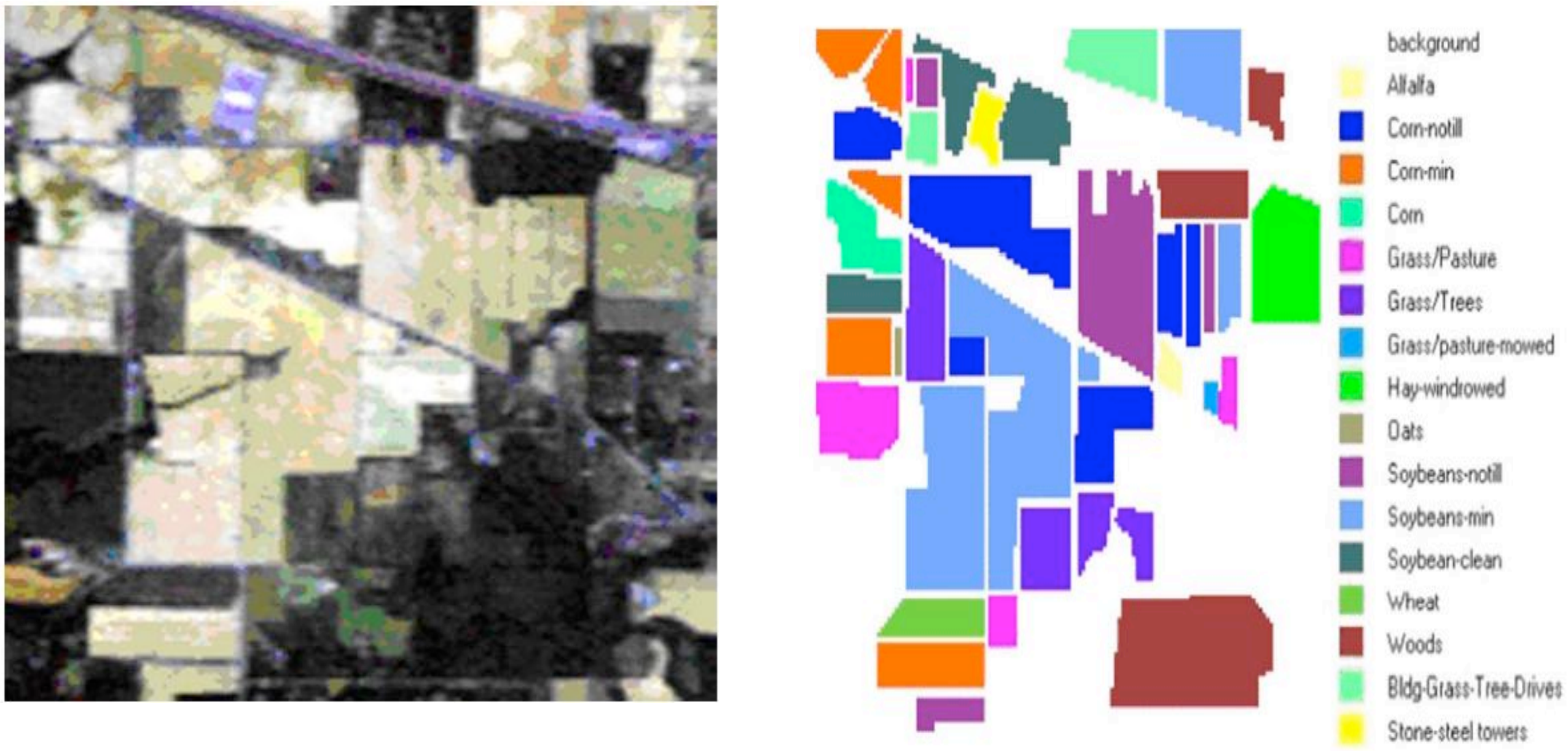

Figure 4. (a) Color composite of the image subset.(b) Ground truth of the area with 16 classes

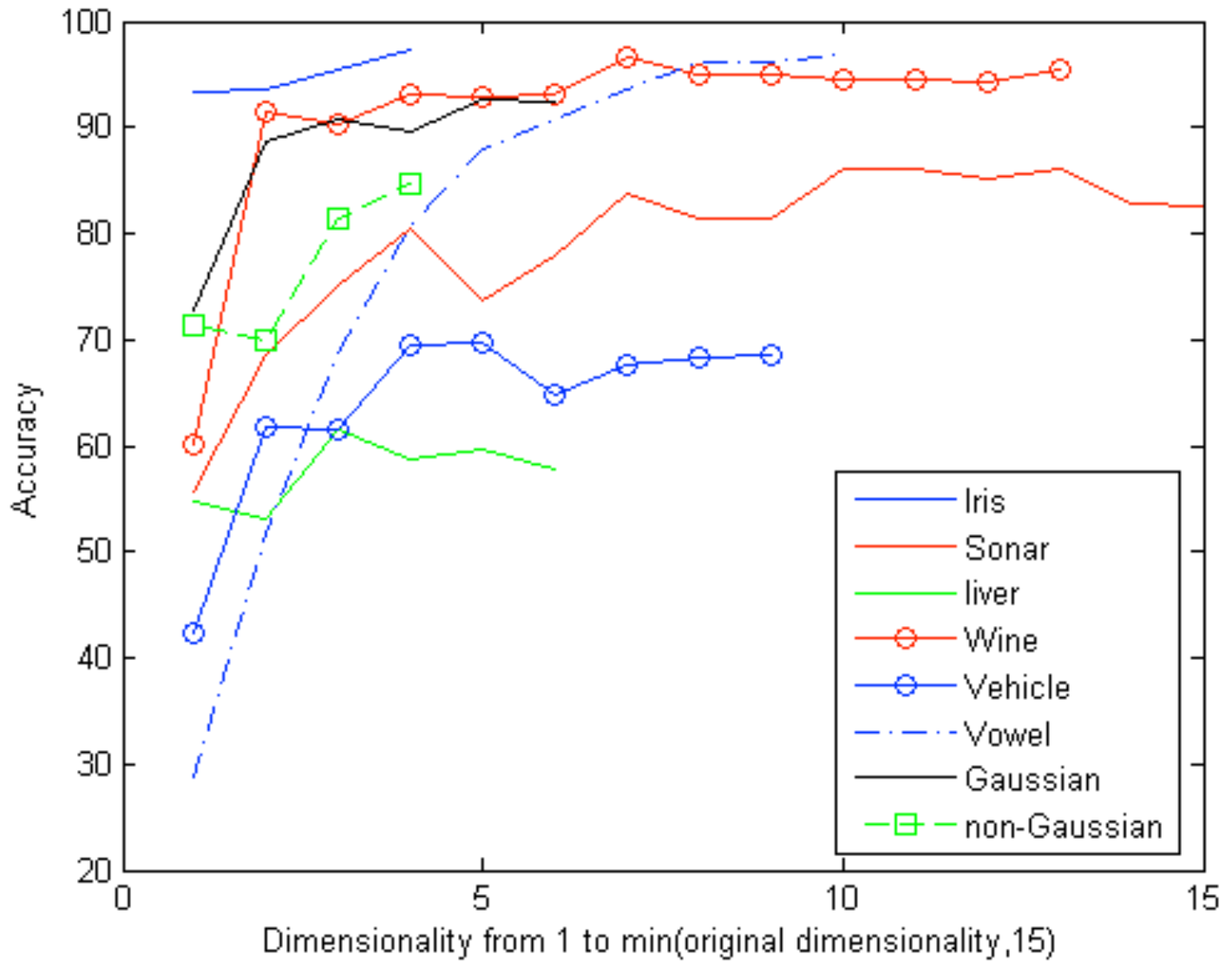

Figure 5. Different accuracies as a function of dimensionality in 6 datasets of UCI, Gaussian and non-Gaussian data

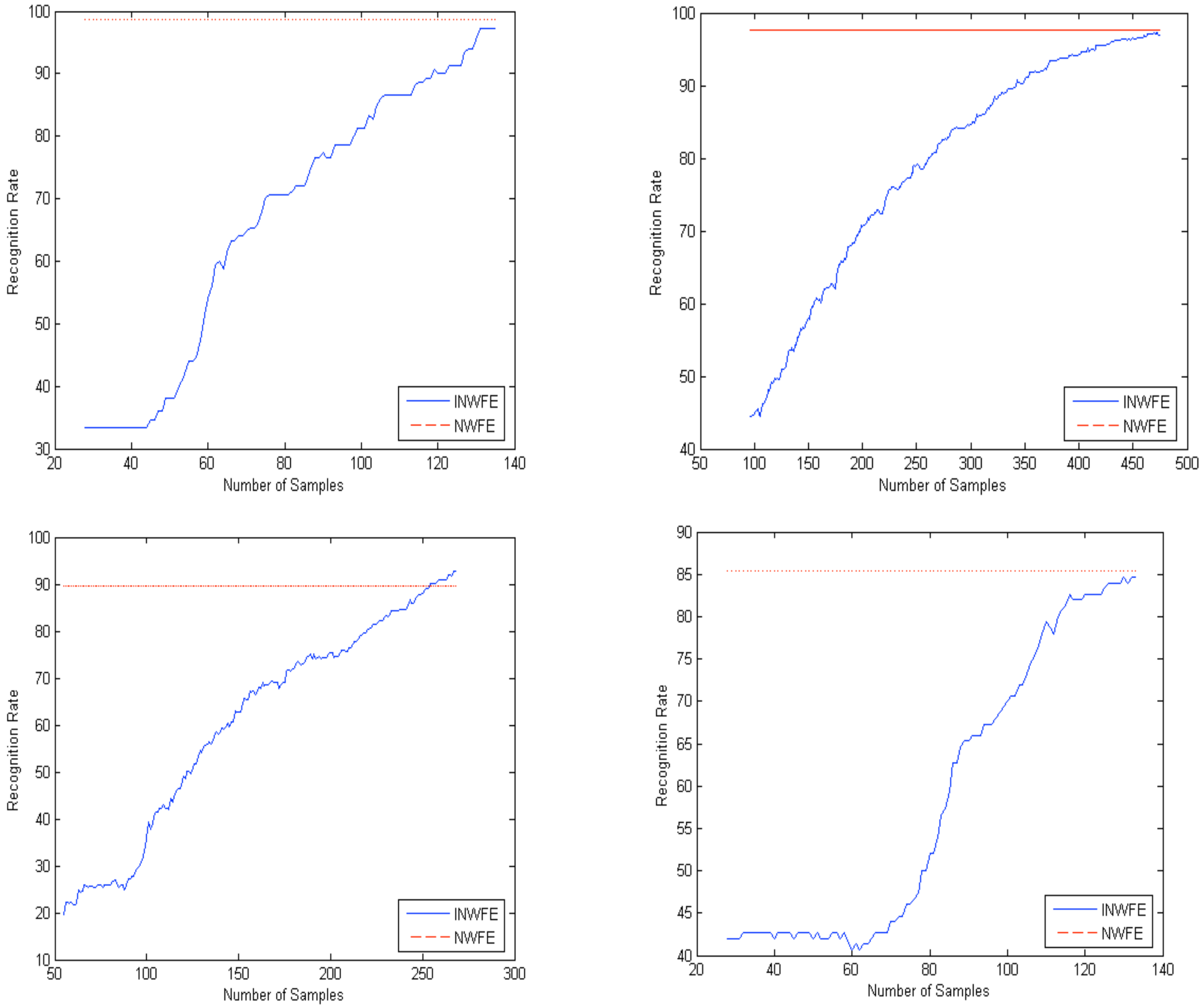

Figure 6. The progress of recognition rate versus number of input samples (a) Iris (b) Vowel (UCI) (c) Gaussian (d) Non-Gaussian data

## References

[1] A.R. Webb, K.D. Copsey,Statistical Pattern Recognition, Third ed., John Wiley & Sons, Chichester, 2011.

[2] K. Fukunaga, Introduction to Statistical Pattern Recognition. Second ed. Academic Press, 1990.

[3] J. H. Friedman, "Regularized discriminant analysis," Journal of the American Statistical Association, vol. 84, pp. 165-175, 1989.

[4] Z. Zhang, G. Dai, C. Xu, M. Jordan, Regularized Discriminant Analysis, Ridge Regression and Beyond. Journal of Machine Learning Research. vol. 11, pp. 2199-2228, 2010.

[5] A. M. Martinez, A. C. Kak, "PCA versus LDA," IEEE Trans. Pattern Anal. Machine Intell. vol. 23, pp. 228-233, 2004.

[6] A. Hossein Sahoolizadeh, B. Zargham Heidari, C. Hamid Dehghani, "A New face recognition method using PCA, LDA and neural network," International Journal of Computer Science and Engineering. vol. 2, 2-4, 2008.

[7] I. Dagher, "Incremental PCA-LDA algorith," International Journal of Biometrics and Bioinformatics (IJBB). vol. 4, pp. 97-101, 2010.

[8] X. Wang, X. Tang, "Dual-space linear discriminant analysis for face recognition," Proc. IEEE Conf. Computer Vision and Pattern Recognition, vol. 2 , pp.564–569, 2004.

[9] J. Ye, R. Janardan, C. H. Park, H. Park, "An optimization criterion for generalized discriminant analysis on undersampled problems," IEEE Trans. Pattern Anal. Mach. Intell. vol. 26, pp. 982–994, 2004.

[10] J. Ye, "Comments on the complete characterization of a family of solutions to a generalized Fisher criterion," J. Mach. Learning Res. vol. 9, pp. 517–519, 2008.

[11] H. Park, B.L. Drake, S. Lee, and C.H. Park, "Fast Linear discriminant analysis using QR decomposition and regularization," Technical Report GT-CSE-07-21, 2007.

[12] J. Yea, Q. Lib, "Rapid and brief communication LDA/QR: an efficient and effective dimension reduction algorithm and its theoretical foundation," Pattern Recognition, vol. 37, pp. 851 –854, 2004.

[13] J. A. Richards, J. Xiuping, "Separability measures for multivariate normal spectral class models," Remote Sensing Digital Image Analysis: An Introduction, Springer, New Yok, 2006.

[14] B. C. Kuo, D. A. Landgrebe, "Nonparametric weighted feature extraction for classification," IEEE Transactions on Geoscience and Remote Sensing. 42 (2004), 1096-1105.

[15] G. Yu, Z. Hu, H. Lu, "Robust incremental subspace learning for object tracking," 16th International conference, ICONIP. vol. 5863, pp. 819-828, 2009.

[16] H. Zhao, P. C. Yuen, "Incremental linear discriminant analysis for face recognition," IEEE Trans. Syst., Man, Cybern. B, Cybern. vol.38, pp. 210–221, 2008.

[17] S. Fidler, D. Skocˇcaj, A. Leonardis, "Combining reconstructive and discriminative subspace methods for robust classification and regression by subsampling," IEEE Trans Pattern Anal Mach Intell. vol. 28, pp. 337–350, 2006.

[18] D. Skocˇaj, M. Uray, A. Leonardis, H. Bischof, "Why to combine reconstructive and discriminative information for incremental subspace learning," Chum et al (eds) Proceedings of computer vision winter workshop, Telcˇ, Czech Republic, 2006, pp. 52-57.

[19] H. Abrishami Moghaddam, M. Matinfar, S. M . Sajad Sadough, K. H. Amiri-Zadeh, "Algorithms and networks for accelerated convergence of adaptive LDA," Pattern Recog. vol. 38, pp. 473–483, 2005.

[20] Y. Aliyari Ghassabeh, H. Abrishami Moghaddam, "Adaptive linear discriminant analysis for online feature extraction. Machine Vision and Applications," vol. 24, pp. 777-794, 2013.

[21] J. Ye, Q. Li, H. Xiong, H. Park, R. Janardan, V. Kumar, "IDR/QR: An incremental dimension reduction algorithm via QR decomposition," IEEE transactions on Knowledge and Data Engineering. vol. 17, pp. 1208–1222, 2005.

[22] B. Raducanu, J. Vitria, "Online nonparametric discriminant analysis for incremental subspace learning and recognition," Pattern Anal Applic. vol. 11, pp. 259–268, 2008.

[23] Blake, C, Merz, C.: UCI repository of machine learning databases. http://www.ics.uci.edu/ mlearn/MLRepositor. html (1998)

[24] [Online]. Available: ftp://ftp.ecn.purdue.edu/biehl/MultiSpec/

[25] B. Mojaradi, H. Abrishami Moghaddam, M. J. Valadan Zoej, R. P. W. Duin, "Dimensionality reduction of hyperspectral data via spectral feature extraction," IEEE Transactions on Geoscience and Remote Sensing, vol. 47, pp. 2091-2105, 2009.

[26] P. K. Varshney, M. K, Arora, Advanced image processing techniques for remotely sensed hyperspectral data, Springer-Verlag, Berlin, Germany, 2004.